\documentclass[10pt,twocolumn,letterpaper]{article}

\usepackage[final]{iccv}      

\definecolor{iccvblue}{rgb}{0.21,0.49,0.74}
\usepackage[pagebackref,breaklinks,colorlinks,allcolors=iccvblue]{hyperref}

\def\paperID{1} 
\def\confName{ICCV}
\def\confYear{2025}

\title{Semantic Aware Feature Extraction for Enhanced 3D Reconstruction}

\author{Ronald Nap \\
Department of Driving Assistance Systems \\
Valeo, San Mateo, USA\\
{\tt\small rnap@berkeley.edu}
\and
Andy Xiao\\
Department of Driving Assistance Systems\\
Valeo, San Mateo, USA\\
{\tt\small andyxiao@valeo.com}
}

\begin{document}
\maketitle
\begin{abstract}

Feature matching is a fundamental problem in computer vision with wide-ranging applications, including simultaneous localization and mapping (SLAM), image stitching, and 3D reconstruction. While recent advances in deep learning have improved keypoint detection and description, most approaches focus primarily on geometric attributes and often neglect higher-level semantic information. This work proposes a semantic-aware feature extraction framework that employs multi-task learning to jointly train keypoint detection, keypoint description, and semantic segmentation. The method is benchmarked against standard feature matching techniques and evaluated in the context of 3D reconstruction. To enhance feature correspondence, a deep matching module is integrated. The system is tested using input from a single monocular fisheye camera mounted on a vehicle and evaluated within a multi-floor parking structure. The proposed approach supports semantic 3D reconstruction with altitude estimation, capturing elevation changes and enabling multi-level mapping. Experimental results demonstrate that the method produces semantically annotated 3D point clouds with improved structural detail and elevation information, underscoring the effectiveness of joint training with semantic cues for more consistent feature matching and enhanced 3D reconstruction.



\end{abstract}    
\section{Introduction}
\label{sec:intro}

Feature correspondence lies at the core of numerous computer vision tasks, serving as the key mechanism through which multiple images of a scene can be related to one another. However, establishing these correspondences is far from trivial. Challenging factors such as extreme viewpoint changes, lighting variations, and textureless areas can degrade the effectiveness of even the most sophisticated feature detectors and descriptors. Moreover, repetitive structures and non-rigid deformations frequently introduce false matches, leading to geometric errors downstream. Consequently, robust feature matching requires descriptors that are not only discriminative with respect to local geometry but also adaptable to complex transformations in real-world images.

While deep learning has significantly advanced the state of the art in keypoint detection and description, current strategies mainly focus on local geometric cues. Most learning-based pipelines strive to maximize geometric repeatability and descriptor discrimination, yet they rarely incorporate higher-level semantic insights that humans intuitively leverage when recognizing objects or scene elements. This gap is particularly essential in settings where semantic cues offer a more robust signal for correspondence. Motivated by this observation, our work investigates the unexploited potential of integrating semantics and feature extraction into an unified multi-task training pipeline.

\section{RELATED WORK}
In this section, we provide an overview of existing methods for feature matching, discuss recent deep learning approaches, and highlight research efforts involving multi-task learning.

\begin{figure*}[t]
    \centering
    \includegraphics[width=16.6cm]{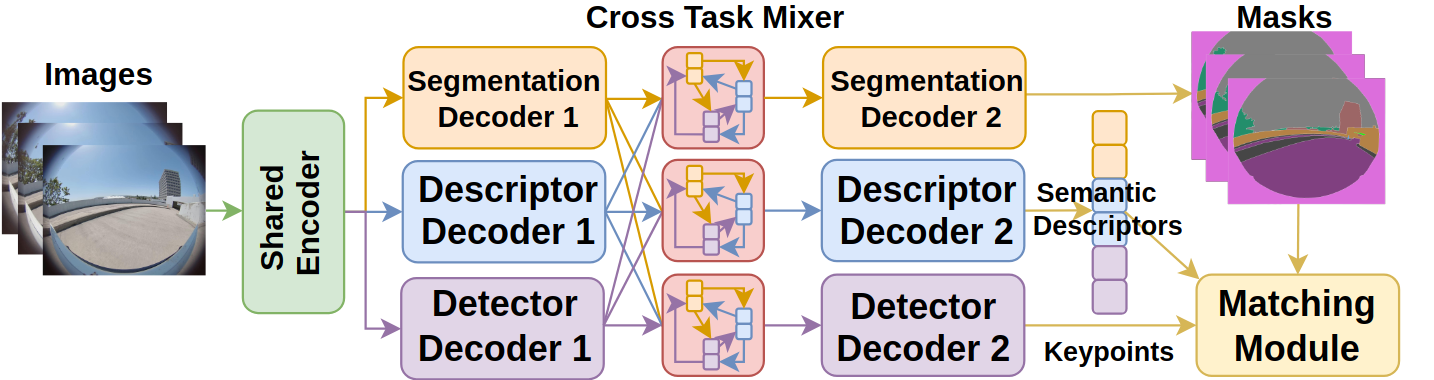} 
    \caption*{Figure 1: We use a single shared encoder to extract features, which pass through an initial decoding stage (Decoder 1) followed by a cross-task mixer module promoting cross-task information exchange. The refined features then undergo a second decoding stage (Decoder 2) to produce segmentation masks, semantic descriptors, and keypoint detections.}
\end{figure*}
\subsection{Traditional Approaches
}

Traditional feature matching methods rely on handcrafted descriptors that encode local image structures within a predefined support region. Classical descriptors such as  SIFT \cite{Lowe2004}, SURF \cite{Bay2008}, ORB \cite{Rublee2011}, BRIEF \cite{Calonder2010} extract distinctive patterns from image patches, facilitating robust keypoint matching across images. These methods incorporate attributes such as scale, rotation, and affine invariance, making them robust to minor viewpoint and illumination changes. However, their reliance on purely geometric properties limits their applicability in challenging scenarios, such as significant viewpoint transformations, textureless regions, or repetitive patterns, where descriptor distinctiveness may degrade. 

\subsection{Deep Learning Approaches}
Deep learning has revolutionized feature matching by replacing handcrafted descriptors with learned representations. Existing methods fall into two categories: sparse keypoint-based and dense descriptor-based approaches. Sparse keypoint-based methods, such as SuperPoint \cite{DeTone2017}, R2D2 \cite{Revaud2019}, and LF-Net \cite{Ono2018}, first detect keypoints and then extract descriptors. While these approaches are computationally efficient, they struggle in textureless or repetitive regions where keypoint detection becomes unreliable.
In contrast, dense descriptor-based methods, such as D2-Net \cite{Dusmanu2019}, ContextDesc \cite{Luo2019}, and Patch2Pix \cite{Zhou2021}, extract features across the entire image, improving matches in low-texture areas. These methods offer higher accuracy but are computationally more expensive. More recently, transformer-based architectures such as LoFTR \cite{Sun2021}, COTR \cite{Jiang2021}, MatchFormer \cite{Wang2022}, and AspanFormer \cite{Chen2022} have leveraged self-attention mechanisms to establish correspondences without explicit keypoint detection. These models excel in textureless and repetitive regions by capturing long-range dependencies. However, despite their improved accuracy, most deep learning-based feature matchers still primarily rely on geometric appearance cues rather than semantic-level reasoning.

\subsection{Multi-task Learning}
Multi-task learning (MTL) seeks to improve generalization by simultaneously training multiple objectives within a single model \cite{Caruana1998, Ruder2017}. By exploiting shared representations across tasks, MTL allows the network to leverage complementary information, often leading to better performance than training each task independently. In computer vision, multi-task models commonly tackle pixel-level tasks such as semantic segmentation, depth estimation, and surface normal prediction under a unified encoder-decoder architecture \cite{Kokkinos2017, Liu2019, Zhang2022}. The central idea is that certain tasks provide mutually beneficial inductive biases—\emph{e.g.}, semantic cues can aid geometric inference, while geometry can refine semantic predictions.
\section{Methodology}
\label{sec:method}

We present a framework that performs keypoint detection, keypoint description, and semantic segmentation using a shared encoder with three specialized decoders. Given an input image $I \in \mathbb{R}^{H \times W \times 3}$, the network produces a keypoint heatmap $H_{\mathrm{kp}} \in [0, 1]^{H \times W}$, a dense descriptor map $D \in \mathbb{R}^{H \times W \times d}$, and a segmentation probability map $\hat{Y} \in [0, 1]^{H \times W \times C}$, where $C$ is the number of semantic classes. In the remainder of this section, we describe the core architectural components of the model, along with the loss functions used during training.

\subsection{Shared Encoder}
Our encoder adopts a U-Net-style design \cite{Ronneberger2015} to capture both local and global context across multiple scales. The downsampling path consists of $N$ convolutional blocks with ReLU activations and pooling layers, progressively reducing the spatial resolution while enlarging the receptive field. At each level $n \in \{1, \dots, N\}$, the encoder produces multi-scale feature maps $\{F^{(1)}, \dots, F^{(N)}\}$ which are then passed forward as skip connections to the corresponding layers in the upsampling path preserving high-frequency spatial details that might otherwise be lost. 

\begin{figure*}[t]
    \centering
    \begin{tabular}{@{}c@{}c@{}}
        \includegraphics[width=0.5\linewidth]{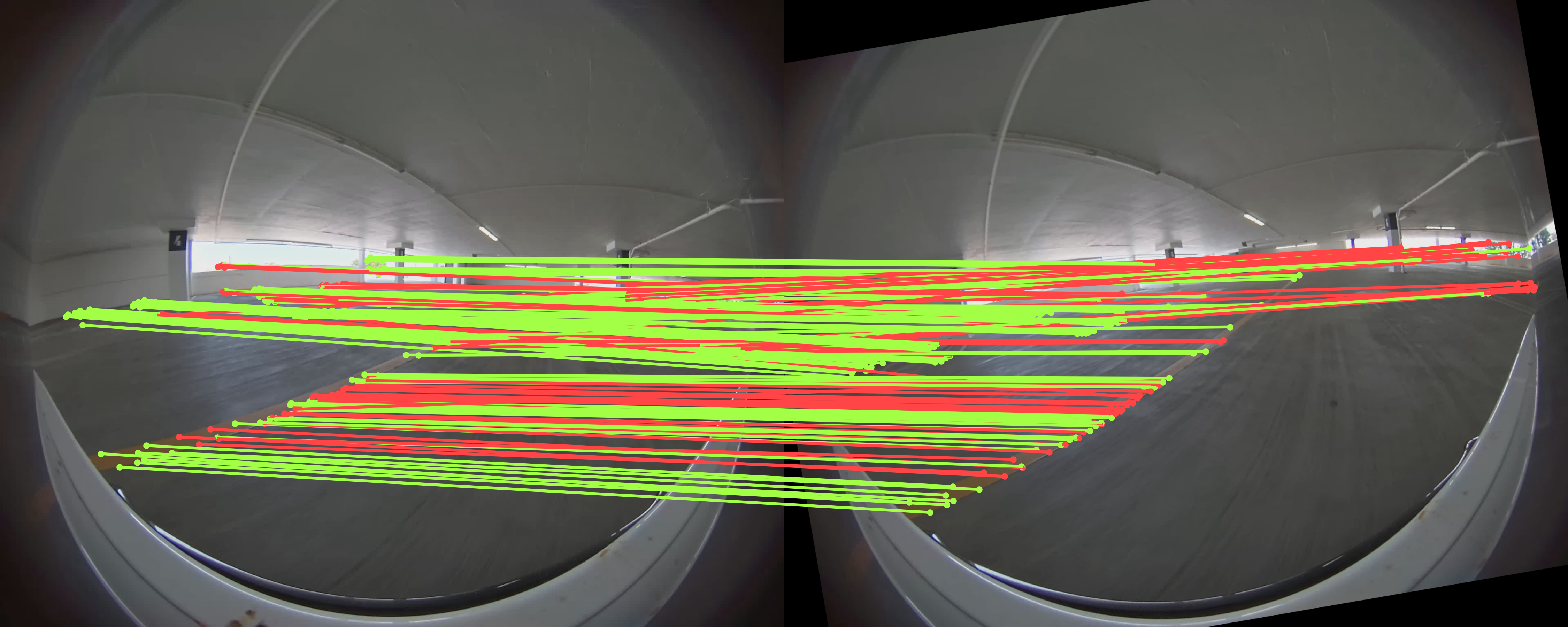} & 
        \includegraphics[width=0.5\linewidth]{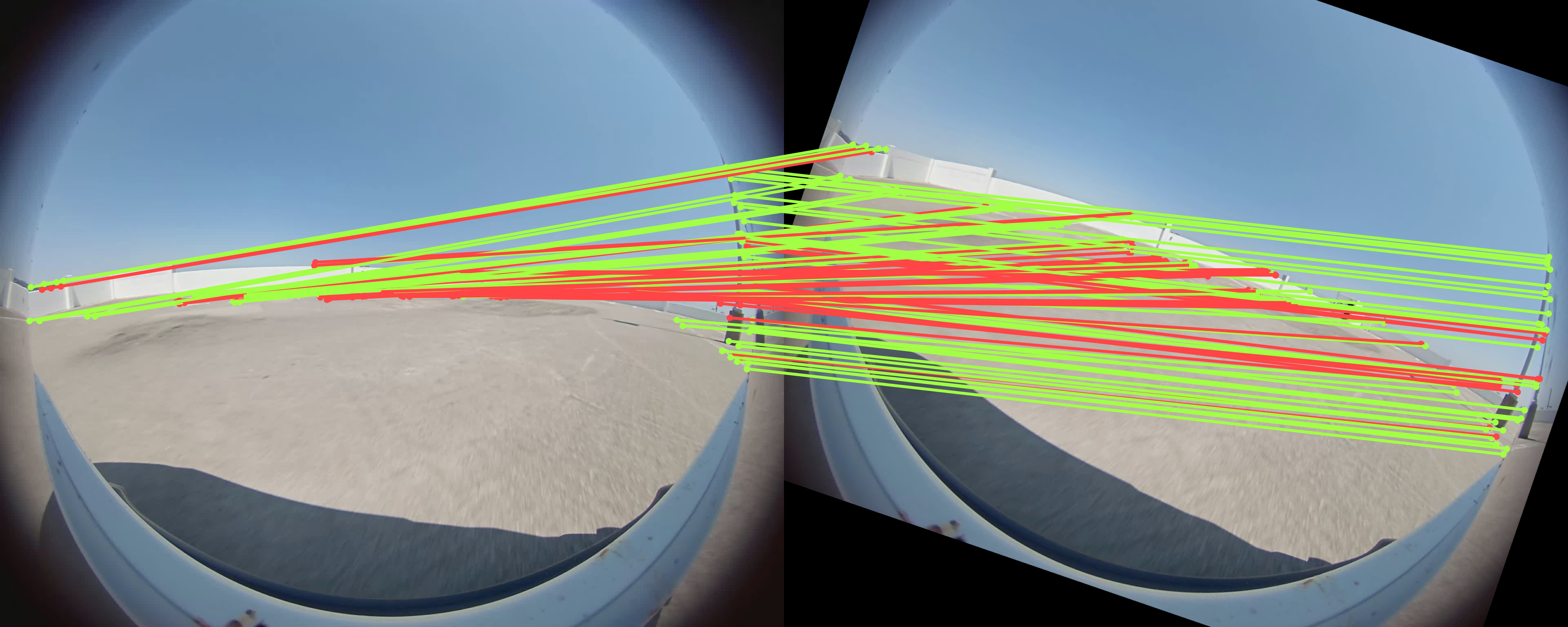} \\
        \includegraphics[width=0.5\linewidth]{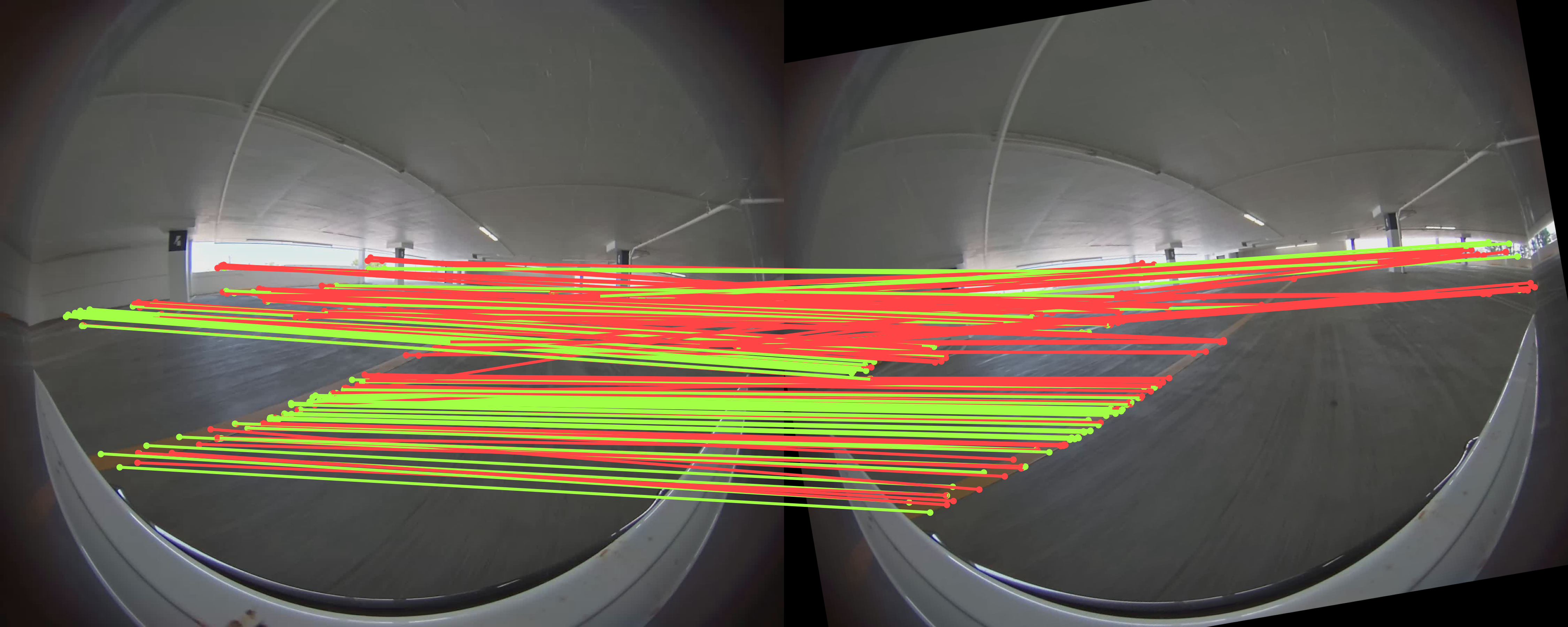} & 
        \includegraphics[width=0.5\linewidth]{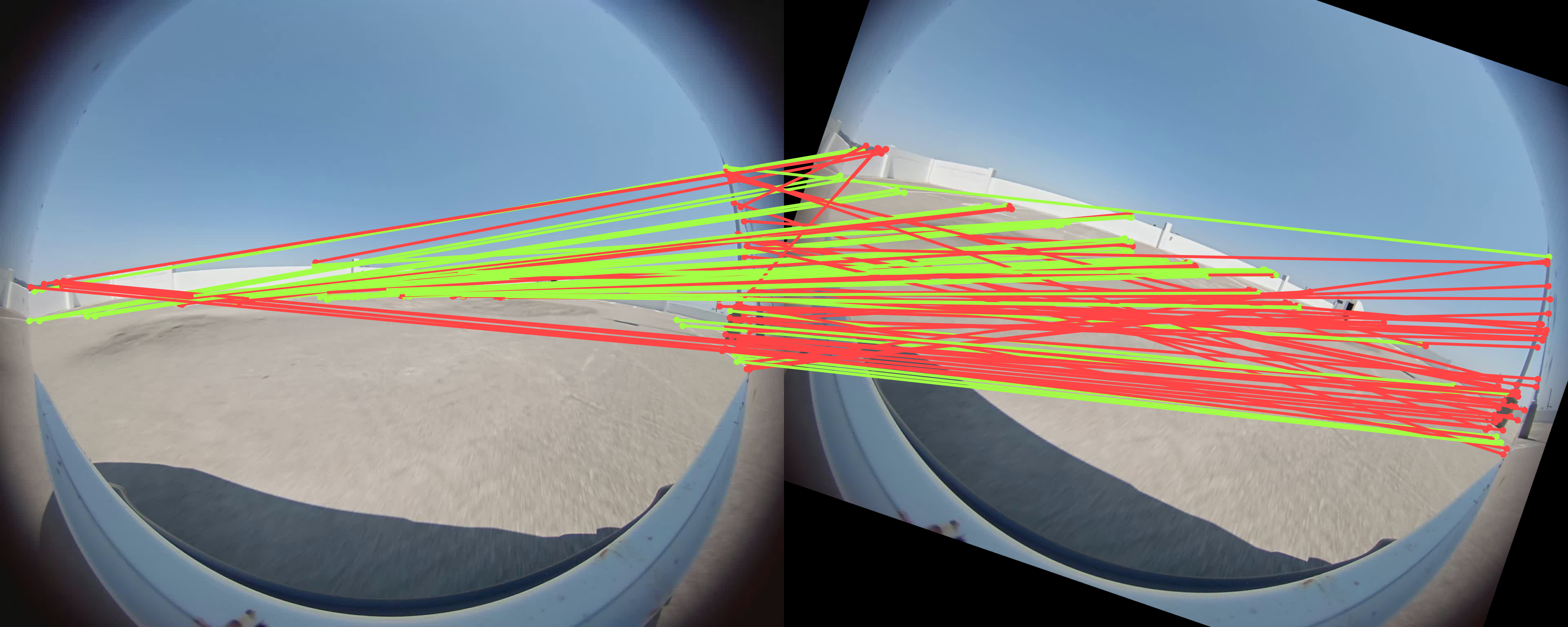} \\
        \includegraphics[width=0.5\linewidth]{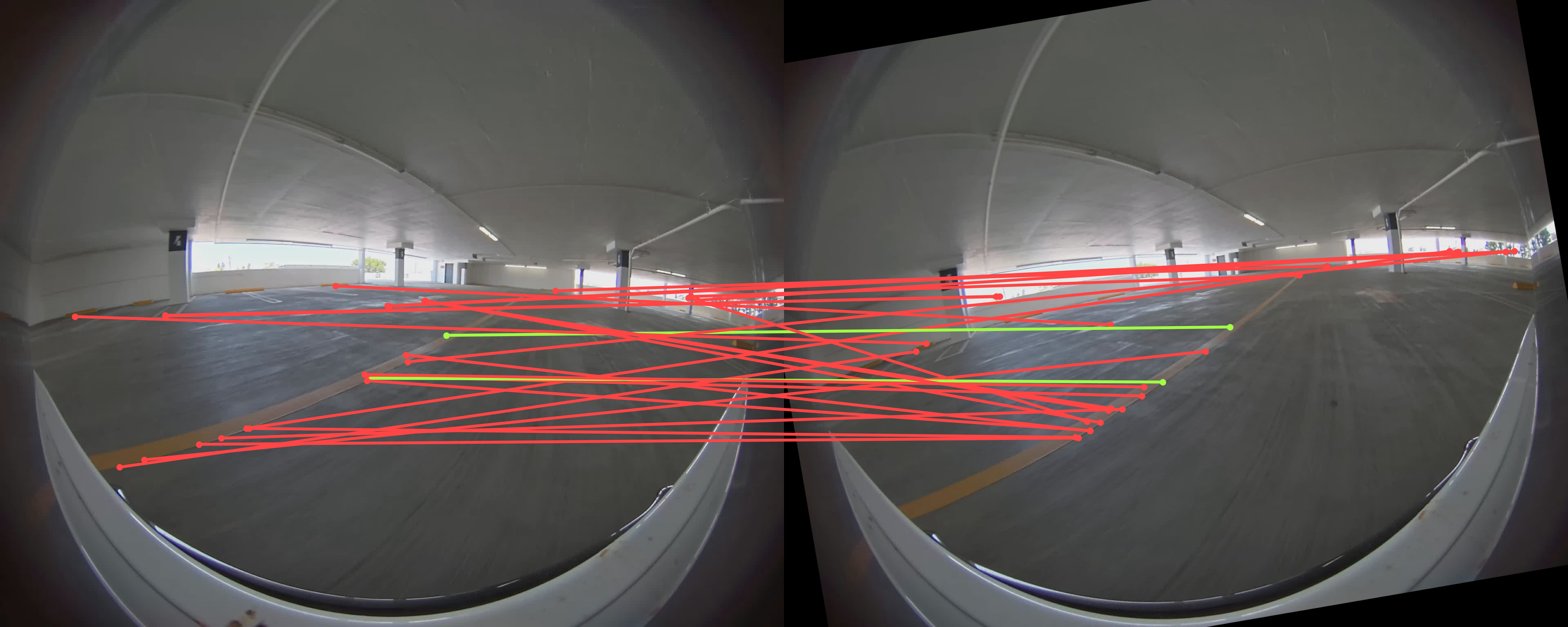} & 
        \includegraphics[width=0.5\linewidth]{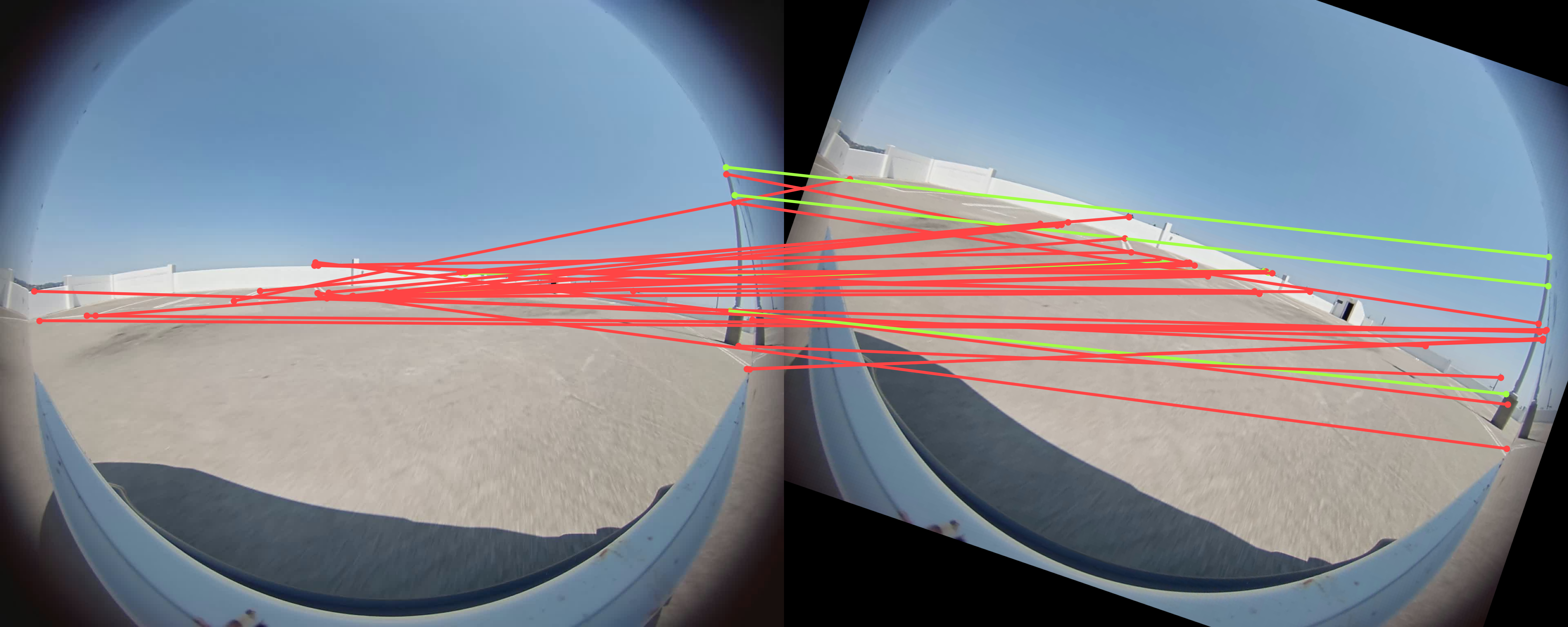} \\
    \end{tabular}
    \caption{Top row: correspondences predicted by our method, middle row: by SuperGlue, and bottom row: by SIFT. Correct matches are shown in green, incorrect in red. A sequential feature matching demo video can be found at \url{https://streamable.com/8vhwcs}}
    \label{fig:matches}
\end{figure*}

\subsection{Cross-Task Mixer} 
\label{sec:ctcm}
Once the encoder produces feature map $\mathbf{z} \in \mathbb{R}^{h \times w \times d_{\mathrm{enc}}}$, we pass $\mathbf{z}$ to each specialized decoder. To effectively integrate knowledge across tasks, we insert a Cross-Task Mixer Module (CTMM) within each decoder. Inspired by existing feature-fusion approaches in multi-task learning \cite{Misra2016, Lopes2023} our CTMM incorporates task-specific embeddings into a unified representation that leverages complementary task interactions. We let $\mathbf{z}_{kp}, \mathbf{z}_{desc}, \mathbf{z}_{seg} \in \mathbb{R}^{d_{\mathrm{task}}}$ 
denote the learned embeddings for each task. The updated feature map is computed as
\begin{equation}
  \tilde{\mathbf{z}}
  \;=\;
  \mathrm{CTCM}\!\bigl(\mathbf{z},\,\mathbf{z}_{kp},\,\mathbf{z}_{desc},\,\mathbf{z}_{seg}\bigr).
\end{equation}
We implement the CTMM as a lightweight attention mechanism that fuses $\mathbf{z}$ with the three task embeddings 
$\{\mathbf{z}_{kp}, \mathbf{z}_{desc}, \mathbf{z}_{seg}\}$. First, we stack these embeddings into a matrix 
$\mathbf{T} \in \mathbb{R}^{3 \times d_{\mathrm{task}}}$. Then, for each spatial location $(u,v)$ of $\mathbf{z}$, we project the local feature $\mathbf{z}_{u,v} \in \mathbb{R}^{d_{\mathrm{enc}}}$ into a query vector $\mathbf{q}_{u,v}$ using a learned weight matrix $W_q \in \mathbb{R}^{d \times d_{\mathrm{enc}}}$:
\begin{equation}
    \mathbf{q}_{u,v} = W_q\,\mathbf{z}_{u,v}, \quad \mathbf{q}_{u,v} \in \mathbb{R}^{d}.
\end{equation}
We then transform the task embeddings $\mathbf{T}$ into key and value matrices via learned projection matrices $W_k, W_v \in \mathbb{R}^{d \times d_{\mathrm{task}}}$:
\begin{equation}
    \mathbf{K} = W_k\,\mathbf{T}, \quad 
    \mathbf{V} = W_v\,\mathbf{T}, \quad \mathbf{K}, \mathbf{V} \in \mathbb{R}^{d \times 3},
\end{equation}
where each column of $\mathbf{K}$ and $\mathbf{V}$ corresponds to one of the tasks $\{\mathrm{kp},\mathrm{desc},\mathrm{seg}\}$. Next, we compute the attention weights 
$\alpha_{u,v} \in \mathbb{R}^{3}$ for each spatial location:
\begin{equation}
    \alpha_{u,v} = 
    \mathrm{softmax}
    \Bigl(
    \tfrac{\mathbf{q}_{u,v}^{\top}\mathbf{K}}{\sqrt{d}}
    \Bigr).
\end{equation}
Finally, the updated feature $\tilde{\mathbf{z}}_{u,v}$ is computed as a weighted combination of the columns in $\mathbf{V}$:
\begin{equation}
    \tilde{\mathbf{z}}_{u,v}
    =
    \sum_{t \in \{\mathrm{kp},\mathrm{desc},\mathrm{seg}\}}
        \alpha_{u,v}[t]\;\mathbf{v}_{t},   
    \quad
    \tilde{\mathbf{z}} \in \mathbb{R}^{h \times w \times d},
\end{equation}
where $\mathbf{v}_{t}$ is the column of $\mathbf{V}$ corresponding to task $t$. Gathering these updated vectors across all spatial locations $(u,v)$ yields the final fused map $\tilde{\mathbf{z}}$, which contains task-aware cues for all three decoders.

\subsection{Multi-task Loss}
Each of the three tasks—keypoint detection, keypoint description, and semantic segmentation—uses a specialized decoder that operates on their associated fused feature map $\tilde{\mathbf{z}}$. The keypoint decoder upsamples $\tilde{\mathbf{z}}$ to the full resolution $(H \times W)$ and applies a $1 \times 1$ convolution plus a sigmoid activation, yielding a probability map $H_{kp} \in [0,1]^{H \times W}$ and is learned via a binary cross-entropy loss:
\begin{align}
\mathcal{L}_{kp} 
=\;& 
-\sum_{u=1}^{H}\sum_{v=1}^{W} 
    \Bigl[
      g_{uv}\,\log\!\bigl(H_{kp}[u,v]\bigr)\nonumber\\
&\quad +\;
      \bigl(1 - g_{uv}\bigr)\,\log\!\bigl(1 - H_{kp}[u,v]\bigr)
    \Bigr],
\end{align}

\begin{figure*}[t]
    \centering
    \includegraphics[width=\linewidth]{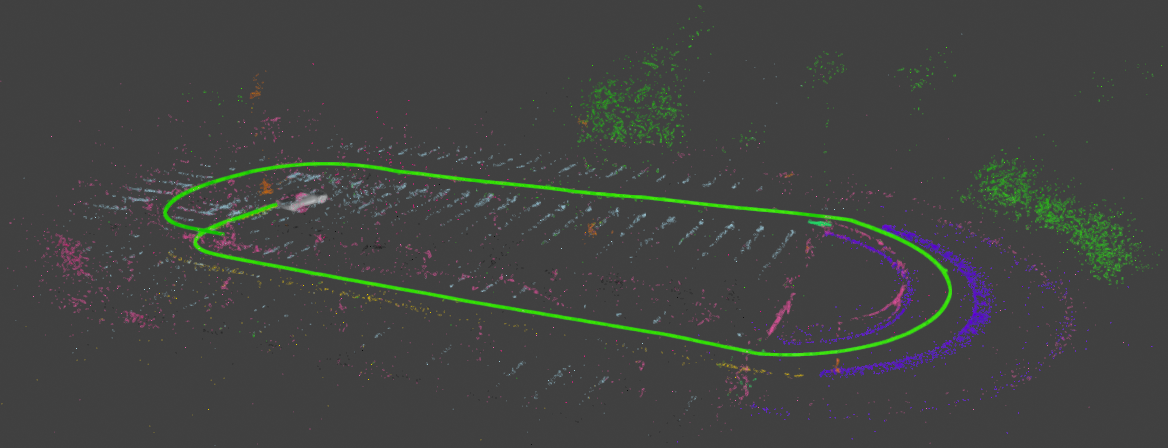}
    \caption{Visualization of our semantic 3D reconstruction. Sequential feature matches and semantic masks are integrated into COLMAP, resulting in a colorized 3D reconstruction. A demonstration video is available at \url{https://streamable.com/jv66z5}.}
    \label{fig:semanticsfm}
\end{figure*}

The descriptor decoder outputs a dense feature map $\mathcal{D} \in \mathbb{R}^{H \times W \times d}$ normalized to unit length per spatial location. We adopt a contrastive margin-based loss \cite{Schroff2015} that encourages similar descriptors for positive pairs $(i,j)\in\mathcal{P}$ and dissimilarity for negative pairs $(i,n)\in\mathcal{N}$:
\begin{align}
    \mathcal{L}_{desc} 
    \;=\;& 
    \sum_{(i,j)\in\mathcal{P}}
    \max\!\Bigl(0,\,
                m - \text{sim}\bigl(\mathbf{d}_i,\mathbf{d}_j\bigr)
           \Bigr)\nonumber\\[6pt]
    &\quad +\;
    \sum_{(i,n)\in\mathcal{N}}
    \max\!\Bigl(0,\,
                \text{sim}\bigl(\mathbf{d}_i,\mathbf{d}_n\bigr) \;-\; \alpha
           \Bigr),
\end{align}
where $\mathrm{sim}(\cdot,\cdot)$ is the cosine similarity, and $m,\alpha$ are margin parameters. The segmentation decoder upsamples $\tilde{\mathbf{z}}$ to predict per-pixel class probabilities $\hat{Y} \in [0,1]^{H \times W \times C}$, trained via the standard multi-class cross-entropy:
\begin{equation}
    \mathcal{L}_{seg} 
    = 
    -\sum_{u=1}^{H}\sum_{v=1}^{W}\sum_{c=1}^{C}
        y_{u,v,c}\,\log\!\bigl(\hat{Y}[u,v,c]\bigr),
\end{equation}
where $y_{u,v,c}\in\{0,1\}$ is the one-hot ground-truth at pixel $(u,v)$ for class $c$. The overall multi-task objective combines these three losses with weighting coefficients $\alpha,\beta,\gamma$:
\begin{equation}
    \mathcal{L}_{\text{total}}
    \;=\;
    \alpha\,\mathcal{L}_{kp}
    \;+\;
    \beta\,\mathcal{L}_{desc}
    \;+\;
    \gamma\,\mathcal{L}_{seg}
\end{equation}
We train the entire network end-to-end using AdamW with a base learning rate of $2\times10^{-4}$ for 100 epochs and a batch size of 16. A short warm-up phase (20 epochs) freezes the early encoder layers, stabilizing low-level features, after which parameters are fine-tuned jointly.
\section{Results}

\subsection{Dataset}
All experiments were conducted in real-world urban parking environments using a test vehicle equipped with a surround-view camera system consisting of four wide-angle fisheye cameras mounted at the front, rear, left, and right sides. Each fisheye camera features a 3-megapixel sensor and captures images at a resolution of 1920×1536 pixels. All sensors were calibrated offline prior to data collection to ensure geometric accuracy. The vehicle was also equipped with GPS-RTK for precise positioning. The parking structure consists of five floors; however, due to poor GPS-RTK reception on the lower levels, data collection was limited to the fifth floor and the upper half of the fourth floor, where positioning accuracy remained reliable. For this study, only images from the front-mounted fisheye camera were used.

The resulting dataset comprises approximately 2,500 images captured under both outdoor and indoor lighting conditions. To support semantic segmentation tasks, a subset of 500 images was manually annotated at the pixel level, covering 19 semantic categories, including signs, walls, buildings, parking lanes, lane dividers, trees, and other relevant urban features. Ground-truth keypoints for feature matching evaluation were generated using a combination of edge detection and region-based techniques. While this dataset is moderate in size for urban parking scenarios and sufficient for initial validation and development, expanding it with additional annotated images from diverse parking environments could further enhance model robustness and generalization.

To introduce geometric variations, the images were subjected to synthetic transformations (see Figure 3), including random rotations (±30°), translations (up to ±15 pixels), and scaling (between 0.8× and 1.2×). Homography-based warps were applied to simulate perspective distortions on planar surfaces such as parking lines and lane dividers, while nonplanar objects like vehicles and pedestrians were excluded to maintain transformation consistency.

\begin{figure}[!htbp]
    \centering
    \begin{subfigure}{\linewidth}
        \centering
        \includegraphics[width=8.3cm]{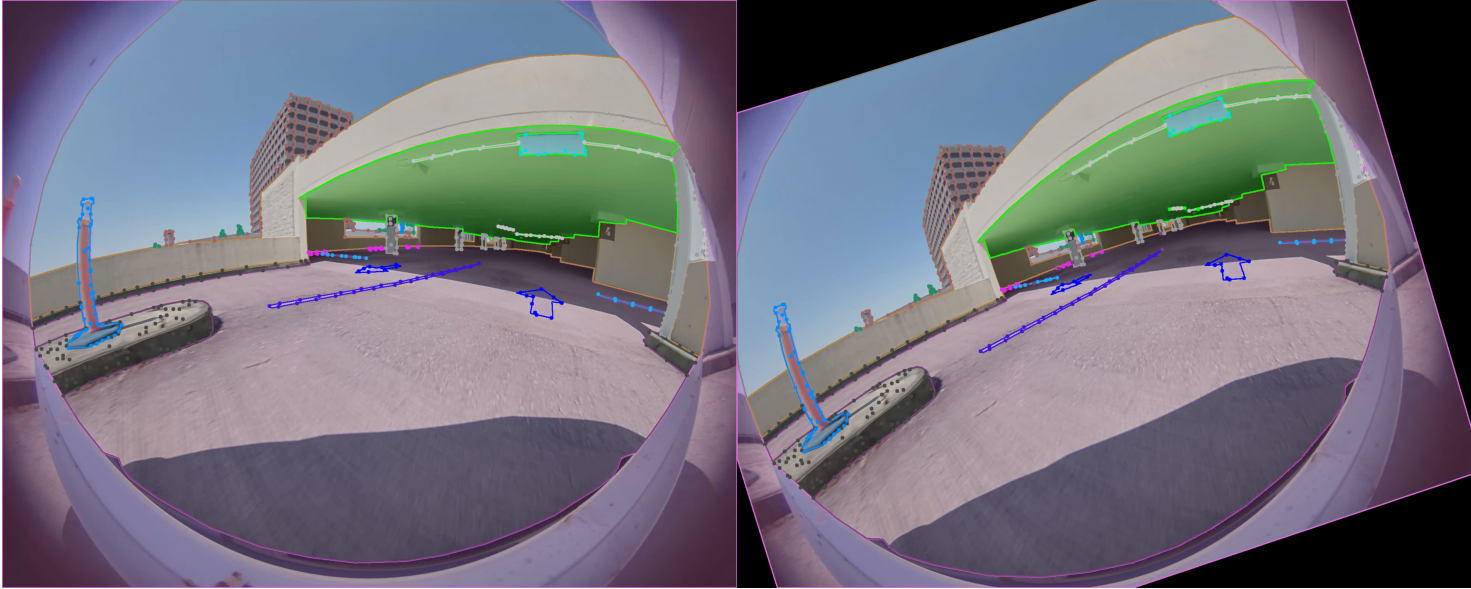}
        \caption{Rotation}
    \end{subfigure}
    \begin{subfigure}{\linewidth}
        \centering
        \includegraphics[width=8.3cm]{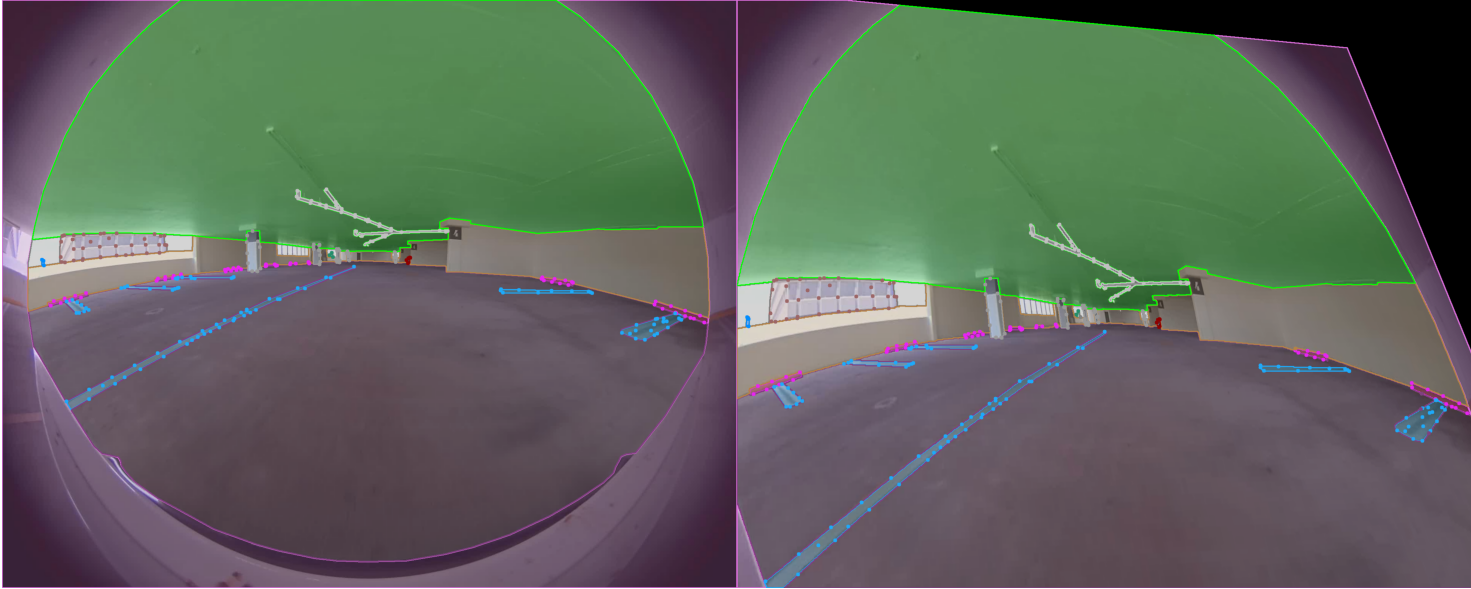}
        \caption{Homography}
    \end{subfigure}

    \caption{Top: Rotation. Bottom: Homography.}
    \label{fig:transform}
\end{figure}

\subsection{Feature Matching Performance}
\label{sec:feature-matching}

We compare our proposed method with traditional keypoint matching approaches: SIFT and ORB with nearest-neighbor (NN) matching, and state-of-the-art learning-based techniques, including SuperGlue [20] and OmniGlue [7]. All methods are evaluated using the same input images and corresponding masks to ensure a fair and controlled comparison.

To quantify the quality of feature correspondences, we measure the keypoint recall and inlier ratio, with the results summarized in Table 1. While classical methods like SIFT and ORB yield reasonable detections in high-texture regions, their performance degrades under challenging conditions such as large rotations or projective transformations. Learning-based methods demonstrate improved robustness in these scenarios, showing higher recall but sometimes struggle in repetitive areas, such as the pillars, parking lines. Our method addresses these limitations by incorporating semantic guidance with SuperPoint-style keypoint detection and learned descriptors, yielding a balance between keypoint repeatability and descriptor distinctiveness. In structured environments like parking garages, this combination enables robust feature detection and reliable matching, leading to the highest Keypoint Recall and Inlier ratio among all evaluated methods.

\begin{table}[t]
    \centering
    \resizebox{\linewidth}{!}{
        \begin{tabular}{@{}lcc@{}}
            \toprule
            \textbf{Method} & \textbf{Keypoint Recall (\%)} & \textbf{Inlier Ratio (\%)} \\
            \midrule
            SIFT + NN        & 38.6  & 64.7 \\
            ORB + NN         & 35.2  & 60.3 \\
            SuperGlue   & 71.3  & 79.5 \\
            OmniGlue    & 73.9  & 81.2 \\
            \textbf{Ours} & \textbf{82.5} & \textbf{83.4} \\
            \bottomrule
        \end{tabular}
    }
    \caption{Quantitative comparison of feature matching performance. Our method outperforms both traditional and learning-based approaches in terms of Keypoint Recall and Inlier Ratio, indicating more reliable and consistent correspondences.}
    \label{tab:table1}
\end{table}

\subsection{3D Reconstruction}
\label{sec:3d-recon}
To evaluate the impact of feature matching on map reconstruction, we integrate our correspondences into the COLMAP [21] structure-from-motion pipeline to reconstruct the upper levels of a parking garage. In addition to estimating camera poses and 3D geometry, our method produces a semantic 3D point cloud in which each point is assigned both spatial coordinates and a semantic label. The reconstruction captures detailed structure across multiple floors, including the elevation change along the connecting ramp. Importantly, it maintains robust feature correspondences despite lighting transitions, as the vehicle moves from the sunlit rooftop into the enclosed lower level. This highlights the robustness of our semantic-aware features under both geometric repetition and variable illumination — two prevalent challenges in parking environments.

To assess the geometric accuracy of the reconstruction, we compare the root mean square error (RMSE) between the camera trajectory estimated by COLMAP and the ground-truth trajectory obtained using high-precision GPS-RTK. After aligning the two trajectories, the RMSE is 0.484 meters. Figure 4 illustrates this comparison, showing close alignment between the estimated and reference trajectories, including through the ramp region.
\begin{figure}[t]
    \centering
    \includegraphics[width=8.3cm]{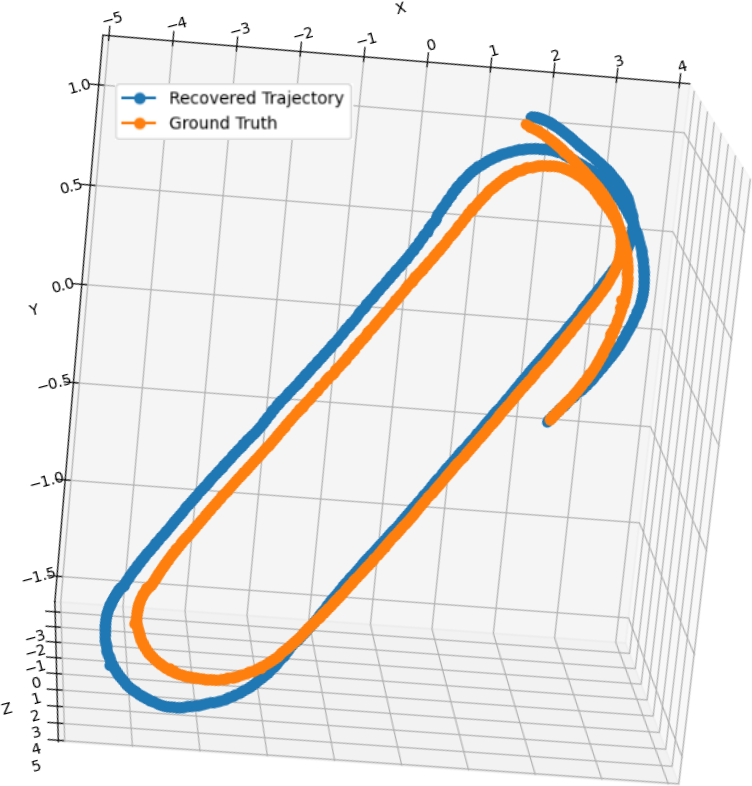}
    \caption{Estimated trajectory vs. ground truth.}
    \label{fig:traj}
\end{figure}

\section{Conclusion}
Our semantic-aware feature extraction strategy addresses common challenges in multi-level parking garages, including repetitive structures and low-texture surfaces. By focusing on stable, consistent features—such as walls and lane markings—and filtering out dynamic or ambiguous elements, the method achieves higher recall and improved geometric consistency. It also supports the detection of elevation changes, such as transitions between floors via ramps. Additionally, the method generates point clouds colorized with predicted semantic labels, providing scene-level understanding that extends beyond geometry alone.
 
Several limitations remain. The reliance on per-pixel semantic predictions can be unreliable in heavily cluttered or occluded areas, and performance tends to degrade on non-planar or highly dynamic surfaces. Future work will focus on strengthening semantic filtering to better handle partial occlusions, expanding the label set for richer scene descriptions, and developing adaptive weighting schemes that more effectively balance feature stability and distinctiveness across diverse indoor–outdoor transitions.
{
    \small
    \bibliographystyle{ieeenat_fullname}
    \bibliography{main}
}

\end{document}